# Robotics in Elderly Healthcare: A Review of 20 Recent Research Projects

Weria Khaksar*

Department of Informatics, University of Oslo, weriak@ifi.uio.no

Department of Department of Mechanical Engineering and Technology Management, Norwegian University of Life Sciences NMBU, weria.khaksar@nmbu.no

Diana Saplacan

Department of Informatics, University of Oslo, dianasa@ifi.uio.no

Lee Andrew Bygrave

Department of Private Law, University of Oslo, l.a.bygrave@jus.uio.no

Jim Torresen

Department of Informatics, University of Oslo, jimtoer@ifi.uio.no

Studies show dramatic increase in elderly population of Western Europe over the next few decades, which will put pressure on healthcare systems. Measures must be taken to meet these social challenges. Healthcare robots investigated to facilitate independent living for elderly. This paper aims to review recent projects in robotics for healthcare (2008 to 2021). We provide an overview of the focus in this area and a roadmap for upcoming research. Our study was initiated with a literature search using three digital databases. Searches were performed for articles, including research projects containing the words "elderly care," "assisted aging," "health monitoring," or "elderly health," and any word including the root word "robot". The resulting 20 recent research projects are described and categorized in this paper. Then, these projects were analyzed using thematic analysis. Our findings can be summarized in common themes: most projects have a strong focus on care robots' functionalities; robots are often seen as products in care settings; there is an emphasis on robots as commercial products; and there is some limited focus on the design and ethical aspects of care robots. The paper concludes with five key points representing a roadmap for future research addressing robotic for elderly people.

**Additional Keywords and Phrases: robotics, elderly, healthcare, literature review, research projects, 2008–2021**

## 1 INTRODUCTION

The relative population of elderly people in Western Europe will continue to increase due to aging, caused by both the postwar baby boom and an increase in life expectancy [1]. By 2050, the working-age population of Europe will be down to 364 million, a 25% reduction compared with the 1995 level. The population aged 65 or older is predicted to increase from 101 million in 1995 to nearly 173 million in 2050 [2]. Thus, the potential support ratio (the number of people aged 15–64 compared to those 65 or older) will decrease drastically, from 4:8 in 1995 to 2:1 in 2050 [3]. The increasing proportion of older people in the population is likely to become one of the most significant social challenges of the 21st century, with implications for nearly all aspects of society, including labor and financial markets, family structures and inter-generational ties, and the demand for goods and services, such as housing, transportation, and social protection [4]. This will result in a decreased number of caregivers for an increased number of care recipients, putting pressure on the quality of our healthcare systems.

The decreasing number of care personnel for elderly people raises the question of how robotics could contribute to maintaining or improving the quality of elderly care. To address these challenges, growing attention is being given to assistive technologies to support seniors in remaining independent and active for as long as possible in their preferred home environment. Robotic systems are among the initiatives offering functionality related to the support of independent living, monitoring and maintenance of personal safety, and enhancement of health and psychological wellbeing through provision of companionship. More generally, the ongoing development of robotics technology in different areas, such as production, transportation, and medicine, shows the importance and potential of robots in improving humans' quality of life. Robots can contribute to healthcare support in terms of capacity (providing medical-related services), quality (accurate and customized performance of specific tasks), cost reduction (support or even replacement of trained personnel), and independence (increased feelings of autonomy and self-management for healthcare users).

This paper aims to review the existing work on the application of robot technology in the healthcare of older adults. The focus is on research projects from 2008 to 2021 that have resulted in different robotic systems being applied in the home care of elderly people. These projects have produced extensive research literature, progress reports, and online descriptions, which are summarized and discussed here. The paper also provides a brief description of ethical issues arising from the implementation of such systems in residential environments. Prior to this study, few survey papers have been published dealing with the application of robots in elderly care since 2008. Studies prior to 2010 are not considered here, since there has been substantial improvement and innovation in the field since they were published. Most studies published since 2010 have focused on psychologically and socially assistive robots [5–8] without looking into healthcare applications.

The concept of ambient assisted living for elderly adults was reviewed in a paper that predominantly considered different sensing tools and corresponding algorithms [9]. Another study was published on the role of healthcare robots at home, with a focus on current issues in the daily activities of older people [10]. However, several of the studied research projects were reported and commercialized after that study's publication. Two other studies have discussed the role of robots in daily activity assistance, with the main focus on general personal care rather than elderly healthcare [11, 12]. Finally, a general review was provided on the application of different robotics and sensory systems in healthcare directed at addressing cognitive, sensory, and motor impairment [13]. Despite the extensive information provided in these papers, the literature in the field lacks a comprehensive review of contemporary projects related to the use of robots in elderly healthcare, emphasizing current progress, main challenges, and future directions. Even though some of the implemented robot systems have been mentioned and briefly discussed previously, a systematic review of current projects is crucial for the future development of robotic systems in the elderly healthcare domain. Research and development in personal care robotics dates back to the 1980s [14]. However, due to rapid technological acceleration, most advances have occurred in the last decade. While the most recent technology builds on earlier attempts to design such systems, all relevant developments in the field are accounted for within the past 10 years' worth of literature.

The main objective of this paper is to summarize the existing robot-based systems in elderly healthcare applications. Existing projects related to the design and implementation of robotic systems for elderly healthcare assistance constitute the primary focus. Details of the proposed technologies and methods are discussed, with an emphasis on the utilized robotic system. Section II introduces the topic of robots for the elderly, and Section III presents the review methodology. In Section IV, the studied projects are introduced, and the findings of the review are presented in Section V. A detailed discussion of the findings is provided in Section VI and concluding remarks and recommendations for further work are presented in Section VII.

## 2 ROBOTICS AND THE ELDERLY

The next subsections provide instructions on how to insert figures, tables, and equations in your document. Although there is no conclusive age at which a person is considered "elderly," retirement age is widely used as the main criterion, which is 60 or 65 years of age in most developed countries. According to the latest statistics, the 60+ age group accounted for 13% of the world's population in 2017. This group represents 25% of the population in Europe and 22% in North America [1]. Figure 1 shows predictions for aging in the world population, and in Europe specifically.

For older people, remaining at home later in life instead of living in an institutional care facility is beneficial in terms of overall quality of life if their needs are adequately addressed. Admission to a nursing home facility has been associated with distress, depression, and emotional problems for patients and their families, as well as early mortality [10]. The application of "aging in place" practices and community care has a number of benefits in terms of cost-effectively meeting the needs of older people and delaying nursing home admission [15], as well as in policy creation. With developing technologies designed to help them retain their independence, older people may be happier and healthier in their old age, provided their needs are met. Many technologies exist at present, including robotic technologies designed specifically to help older people.

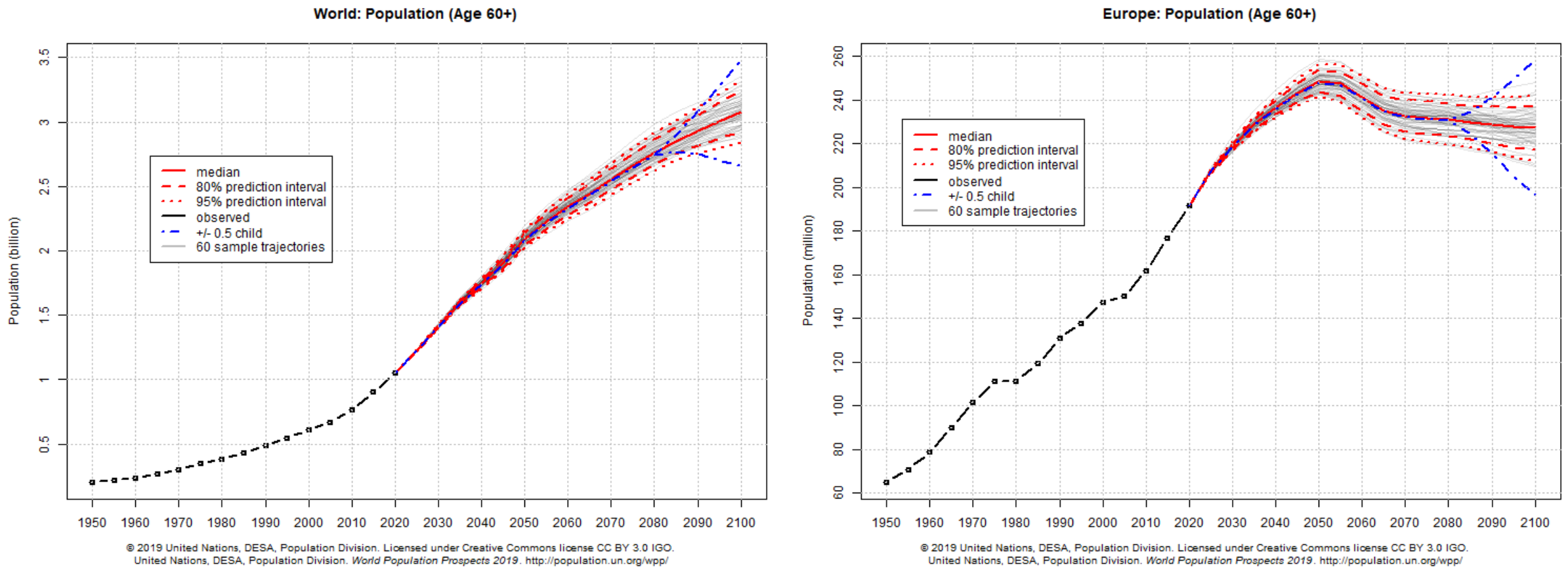

(a) (b)

Figure 1. General prediction of the world population: (a) the world population aged over 60 and (b) the population over 60 years old in Europe. Source: United Nations, Department of Economic and Social Affairs, Population Division (2019). World Population Prospects: The 2019 Revision. https://population.un.org/

The use of robotics for a variety of applications has become increasingly common in recent years, especially in replacing repetitive or potentially dangerous tasks. There have been many recent examples of robotics technology in practice, such as autonomous vehicles, package delivery drones, and robots that work side by side with skilled human workers in factories. One of the most exciting areas in which robotics has tremendous potential to make an impact on our daily lives is healthcare. A healthcare robot has the aim of promoting or monitoring health, assisting with tasks that are difficult to perform due to health problems, or preventing further health decline. Health in this sense encompasses not just physical, but also mental, emotional, and psychosocial dimensions.

Even though “assistance” is the main function of healthcare robots, as the robotic system should assist the user in certain ways, healthcare robots can have many different functions and applications, such as rehabilitation and socialization. Rehabilitation robots are physically assistive devices that are not primarily communicative or perceived as social entities. Their job is to assist a person in physical therapy with instructions and assessments. A social robot is easily understood and likable for older people to interact with and acts as a companion. Social robots can also help a person perform tasks to improve their day-to-day life. Social robots can be further categorized into service robots or companionship robots [16]. Service-type robots are assistive and are designed to support people living independently by assisting with mobility, completing household tasks, and monitoring health and safety [17]. A social robot can exhibit verbal and non-verbal social skills and cues by interacting with humans in a personalized way [18]. Companion robots do not assist the user in performing tasks but aim to improve quality of life by acting as companions. Some robots provide both companionship and assistance. Robots can potentially support people’s independent living and increase their quality of life by addressing functional, psychological, and medical monitoring needs in a comparable way to more simple and less advanced existing assistive technologies (ATs), such as wearables. Many of these robotic systems will be discussed in terms of how they may help address the problems older people encounter in their everyday lives, looking at practical solutions as well as the advantages and disadvantages of robotic technology.

## 3 METHOD

### 3.1 Literature Search Procedure

Tables are “float elements” which should be inserted after their first text reference and have specific styles for identification.  Do not use images to present tables, or they will be inaccessible to readers using assistive technologies.

The main focus was to find relevant research projects on the application of robotic technology in elderly care rather than individual papers describing a single case study or experiment. A general search was undertaken to find relevant project websites. Based on the information gathered from each project, corresponding publications regarding robotic systems for elderly care were searched for on three databases: Elsevier Science Direct, IEEE Xplore, and Google Scholar. Searches were performed for articles including “elderly care,” “assisted aging,” “health monitoring,” and “elderly health,” and any word including the root word “robot.” As previously mentioned, only projects presented in the past 15 years were examined in further detail. Duplicate sources were eliminated, and inclusion criteria were then applied to narrow down the search results. Each project needed to apply a commercial or custom-made robotic system with improved functionality and user studies. It had to have the capability of being controlled entirely by the trial participant or individual user and include safety features, such as speed control and emergency stop. Articles that presented studies evaluating the design and performance of these manipulators were also included. A set of exclusion criteria was applied to help refine the articles included in the review. Articles were not included if they presented a system with functions that fell outside the scope of elderly healthcare. Robots designed merely for entertainment or social purposes, and those that had other purposes within the medical field, such as surgery robots, were not included in the final list. Furthermore, only articles that were directly funded by the projects were considered. Finally, 20 relevant research projects were selected and compiled, as presented in Section IV.

### 3.2 Analysis of elderly care research projects

After the initial screening of the projects, we analyzed each project both individually and in comparison with the other projects. We adopted an inductive approach, following the steps described in the thematic analysis [19], in order to identify the main categories and themes addressed by recent projects on care robots. We proceeded in a systematic manner. Step 1 involved assigning a unique ID for each project. We selected at least two main references describing each of the projects and their main applications and findings, and we familiarized ourselves with each of the projects. In Step 2, we generated codes for each individual project. Authors 1 and 2 labeled each individual project, and the codes generated by each of the authors were merged. Some of these appeared in the codes generated by both authors. In total, 1202 (non-unique) codes were generated together by both authors. All the following steps were carried out by Author 2. In Step 3, the codes were sorted into potential categories, which then formed themes. Forty-seven categories of codes were identified (Figure 2 shows an overview of the categories covered per project). Some of these include robot, elderly, functionality, interaction, care, users, product, ethics and privacy, and safety. In Step 4, we gathered similar categories into potential themes. In the first round, we identified 10 central themes that the projects flagged, including functionality of the robot and care, social- and tele-robots (robots that could be remotely controlled, or that allow users to communicate with others remotely), robots as a product, different users, human–robot interaction, non-controlled settings, ethical aspects, design of the robot, research, and other aspects.

In Step 5, the final step, we revised and redefined the themes that emerged after careful grouping of the initial themes generated in the first round. The final themes that emerged from the review of 20 different projects on robotics in elderly care were: a) care robots’ functionalities; b) robots as products in care settings; c) a strong emphasis on robots as commercial products; and d) the design and ethical aspects of care robots. An overview of the themes, and some examples of categories and sub-categories are illustrated in Figure 3. Finally, the themes were discussed by the authors (Author 1, Author 2). After providing an overview of robotic systems for elderly care in the next section, we present the abovementioned key themes.

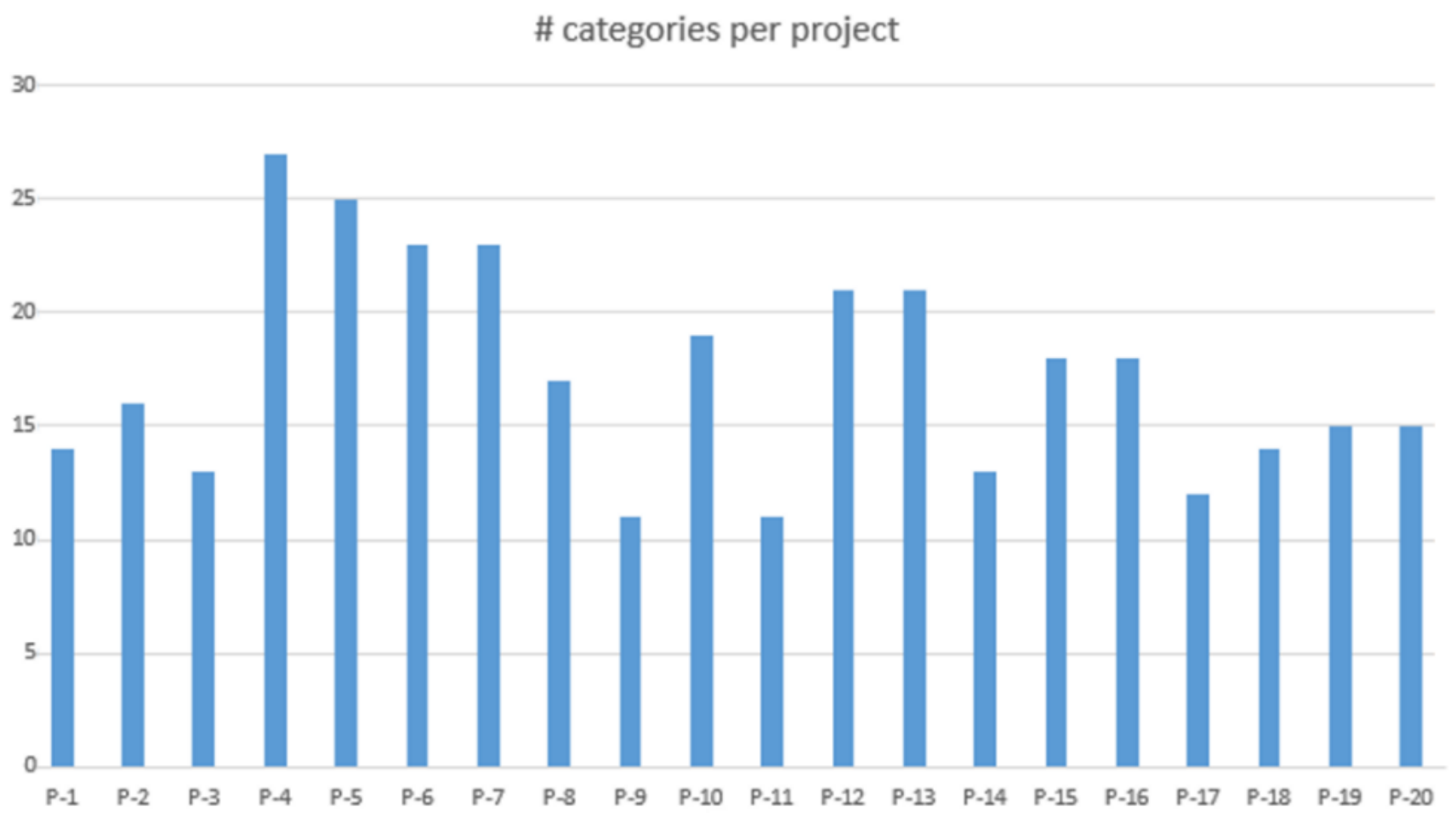

Figure 2. Identified categories per project.

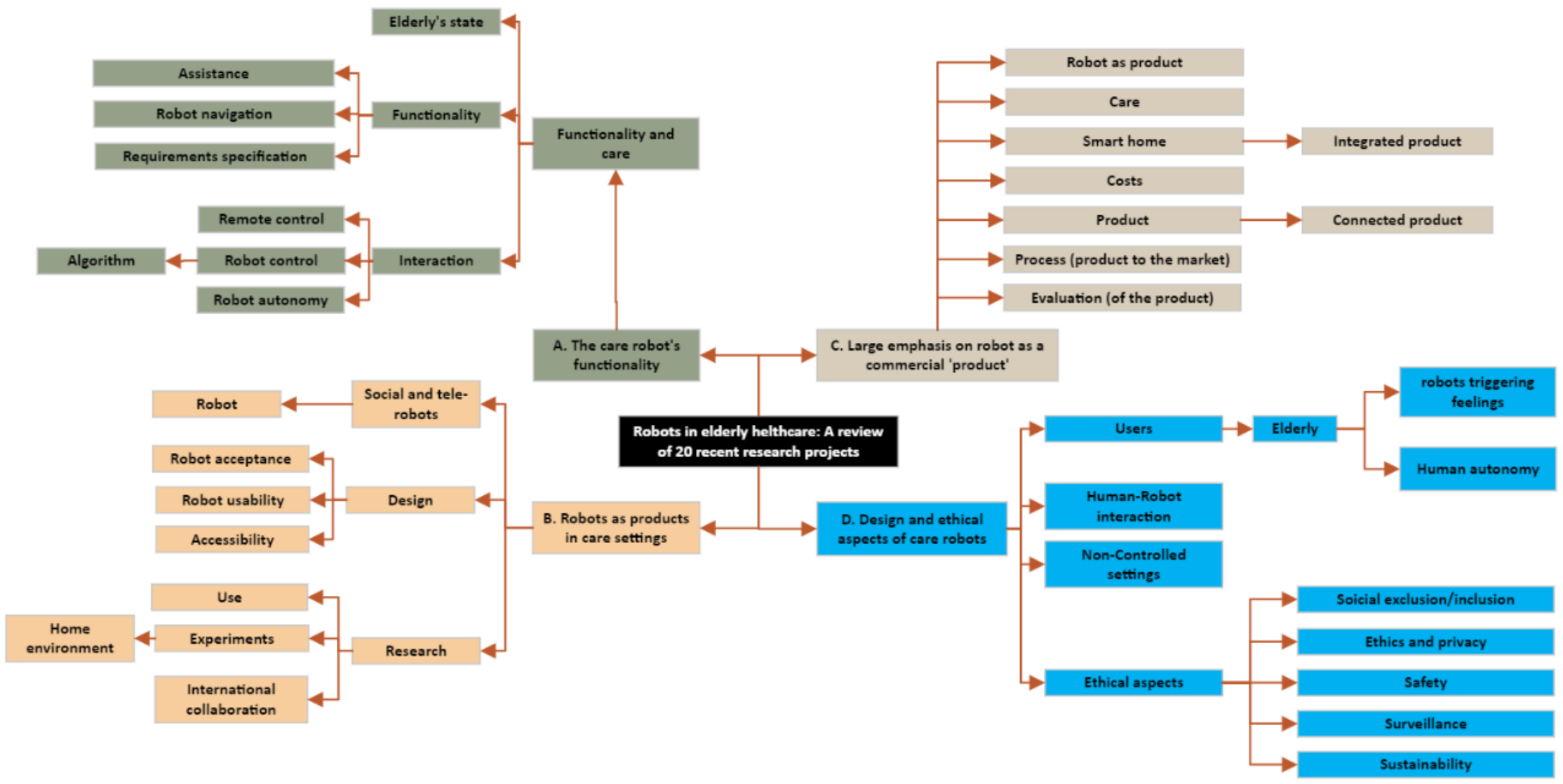

Figure 3. Overview of themes and examples of categories and sub-categories.

## 4 ROBOTIC ELDERLY CARE SYSTEMS (PROJECTS)

In this section, each studied project is introduced and discussed, and the functionality of each system is explained and compared to other projects. A summary of these projects with the corresponding details is provided in Tables 1 and 2.

Table 1. List of studied projects with corresponding robots, sensing devices, and functionalities.

| # | Project | Coordinator | Duration | Budget (Mill. Euros) | Robotic Platform | Physical Description | Implemented Sensors | Functional Capabilities | Interfacing Methods |
|---|---|---|---|---|---|---|---|---|---|
| 1 | CompanionAble [20] | The University of Reading, UK | 2008–2012 | 10.7 | Hector | Mobile, human-like, approx. 1 m tall | Laser scanner, camera, microphone, speaker | Smart home, time management, content generation, reminder function, back tray, house general monitoring | Audio, video, touch screen, eyes showing robot's emotions, telepresence |
| 2 | DOMEO [21] | Robosoft (France) | 2009–2012 | 2.4 | Kompai | Mobile, human-like, approx. 1 m tall | Laser scanner, 3D camera, microphone, speaker | Facial recognition, emotion detection, fall detection, health monitoring, medical data recording, movement support | Audio-visual communication, telepresence, eyes showing robot's emotions |
| 3 | Mobiserv [22] | Stichting Smart Homes, Netherlands | 2009–2013 | 3.6 | Kompai | Mobile, human-like, approx. 1 m tall | Laser scanner, 3D camera, microphone, speaker | Facial recognition, emotion detection, fall detection, health monitoring, medical data recording, movement support | Audio-visual communication, telepresence, eyes showing robot's emotions |
| 4 | ExCITE [23] | Örebro University | 2010–2013 | 2.8 | Giraff | Mobile, human-like, approx. 1 m tall | Laser scanner, camera, microphone, speaker | Time management, content generation, reminder function, object recognition and categorization | Audio-visual communication, telepresence |
| 5 | ACCOMPANY [24] | University of Hertfordshire | 2011–2014 | | Care-O-bot 3 | Mobile, human-like, approx. 1.5 m tall | 3D camera, laser scanner, tactile sensors, microphone, speaker | Smart home, time management, content generation, reminder function, environment and activity monitoring, object recognition and categorization, grasping | Audio, video, touch screen, telepresence |
| 6 | HOBBIT [25] | Technische Universitaet Wien, Autria | 2011–2015 | 3.8 | HOBBIT | Mobile, human-like, approx. 1 m tall | 3D camera, laser scanner, tactile sensors, microphone, speaker | Facial recognition, emotion detection, fall detection, health monitoring, medical data recording, movement support | Audio, video, touch screen, telepresence |
| 7 | Giraff Pluss [26] | Örebro University | 2012–2015 | 4.1 | Giraff | Mobile, human-like, approx. 1 m tall | Laser scanner, camera, microphone, speaker | Time management, content generation, reminder function, object recognition and categorization | Audio-visual communication, telepresence |
| 8 | Robot_Era [27] | Scuola Superiore Sant'Anna (SSSA), Pisa, Italy | 2012–2015 | 8.5 | CoRo, DoRo, and ORo Robots | Mobile, Human-like, approx. 1.5 m tall | Laser scanner, camera, microphone, speaker | Facial recognition, emotion detection, fall detection, health monitoring, medical data recording | Audio, video, touch screen, telepresence |
| 9 | SILVER [28] | The Technology Strategy Board, UK | 2012–2016 | 4.2 | LEA | Mobile stroller, approx. 1 m tall | Laser scanner, camera, microphone, speaker | Facial recognition, emotion detection, fall detection, health monitoring, medical data recording, movement support | Audio, video, touch screen, telepresence |
| 10 | VictoryaHome [29] | Smart Homes, Netherlands Norwegian University of Science and Technology | 2013–2016 | 2.4 | Giraff | Mobile, human-like, approx. 1 m tall | Laser scanner, camera, microphone, speaker | Time management, content generation, reminder function, object recognition and categorization | Audio-visual communication, telepresence |
| 11 | TERESA [30] | University of Oxford | 2013–2018 | 3.9 | Giraff | Mobile, human-like, approx. 1 m tall | Laser scanner, camera, microphone, speaker | Time management, content generation, reminder function, object recognition and categorization | Audio-visual communication, telepresence |
| 12 | RAMCIP [31] | Ethniko Kentro Erevnas Kai Technologikis Anaptyxis, Greece | 2015–2017 | 4.0 | RAMCIP | Mobile, human-like, approx. 1 m tall | Laser scanner, camera, microphone, speaker | Facial recognition, emotion detection, health monitoring | Audio, video, touch screen, telepresence |
| 13 | ENRICHME [32] | Althea Italia SPA, Italy | 2015–2018 | 4.0 | Kompai | Mobile, human-like, approx. 1 m tall | Laser scanner, 3D camera, microphone, speaker | Facial recognition, emotion detection, fall detection, health monitoring, medical data recording, movement support | Audio-visual communication, telepresence, eyes showing robot's emotions |
| 14 | GrowMeUp [33] | University of Coimbra, Portugal | 2015–2018 | 3.4 | Hugo | Mobile, Human-like, approx. 1.5 m tall | Laser scanner, camera, voice, temperature sensor | Time management, content generation, reminder function, back tray, house general monitoring | Audio-visual communication |
| 15 | Mario [34] | National University of Ireland Galway, Ireland | 2015–2018 | 4.0 | Kompai | Mobile, human-like, approx. 1 m tall | Laser scanner, 3D camera, microphone, speaker | Facial recognition, emotion detection, fall detection, health monitoring, medical data recording, movement support | Audio-visual communication, telepresence, eyes showing robot's emotions |
| 16 | RADIO [35] | National Center for Scientific Research, Greece | 2015–2018 | 3.8 | Zacharias | Mobile, human-like, approx. 1 m tall | Laser scanner, camera, microphone, speaker | Time management, content generation, reminder function, object recognition and categorization | Audio-visual communication, telepresence |
| 17 | SocialRobot [36] | University De Coimbra, Portugal | 2011–2015 | 1.0 | SocialRobot | Mobile, human-like, approx. 1 m tall | Laser scanner, camera, microphone, speaker | Facial recognition, emotion detection, health monitoring, data recording | Audio-visual communication, telepresence, eyes |
| 18 | MECS [37] | University of Oslo, Norway | 2016–2021 | 1.2 | Sony Aibo, NAO, TurtleBot2, Fetch, Neato, Pepper | NAO and TurtleBot: mobile, human-like, approx. 0.75 m tall; Aibo: companion robot, 0.2 m; Neato, mobile, service robot, 0.15 m | Implemented sensors NAO: speaker; voice TurtleBot2: LiDar, camera, thermal camera, radar-based sensors Neato: InfraRed sensor; speaker | Facial recognition, emotion detection, fall detection, health monitoring, medical data recording | Audio-visual communication |
| 19 | VIROS [38] | University of Oslo, Norway | 2019–2023 | 2.5 | TIAGo | Human-like, with one robot arm manipulator, 1.1–1.45 m tall | RGB camera, infrared camera, speaker, microphone, LiDar, | Sensing and control of the arm manipulator; TIAGo's own sensors; thermal and depth camera | Non-verbal communication |
| 20 | HIRo [39] | Institute for Energy Technology, Sunnaas Hospital, Halodi Robotics; Norway | 2020–2024 | 0.5 | EVE | Mobile, human-like; approx. 1.83 m tall | Camera, Microsoft Azure Kinect with depth and color cameras, accelerometers in arms, torso and leg | Facial recognition, emotion detection, health monitoring, medical data recording, movement support, healthcare assistance in hospitals | Audio-visual communication, telepresence, eyes showing robot's emotions |

Table 2. Taxonomy of the projects based on the main activities in elderly healthcare.

| # | Activity | Description (How) | Projects 1 | 2 | 3 | 4 | 5 | 6 | 7 | 8 | 9 | 10 | 11 | 12 | 13 | 14 | 15 | 16 | 17 | 18 | 19 | 20 |
|---|---|---|---|---|---|---|---|---|---|---|---|---|---|---|---|---|---|---|---|---|---|---|
| 1 | Companionship | Any kind of hardware or software creation designed to give companionship to a person | ● | ● | ● | ● | | ● | ● | ● | ● | ● | ● | ● | ● | ● | ● | ● | ● | ● | | ● |
| 2 | Time and activity management | Keeping track of user's daily activities and times and sending corresponding reminders to the user | ● | ● | ● | ● | | ● | ● | ● | | ● | ● | | | ● | ● | ● | | | | |
| 3 | Smart home | Equipping the user's home with mounted, on-body, and on-robot sensing devices and combining them in a single system | ● | ● | | ● | | ● | ● | ● | | | ● | ● | ● | | ● | ● | | | | |
| 4 | Physical rehabilitation | Helping the elderly user with illness or injury to restore lost skills and regain maximum self-sufficiency | | | | | ● | | | | | | | | | | | | | | | ● |
| 5 | Mental therapy | Engaging with the user to improve mental illness through interactive communication and activities | ● | ● | ● | ● | ● | ● | ● | ● | ● | ● | ● | ● | | ● | ● | | ● | ● | | ● |
| 6 | Communication | Enabling the elderly user to communicate virtually with family members, friends, caregivers, and medical officers | ● | ● | ● | ● | | ● | ● | ● | ● | ● | ● | ● | ● | ● | ● | ● | ● | ● | ● | ● |
| 7 | Housework | Performing physical housework, such as moving or repositioning household items | | | | ● | | | | | | | | | | | | | | | ● | |
| 8 | Health monitoring | Monitoring the health of the elderly user by utilizing various sensory equipment, such as cameras | ● | ● | | ● | | ● | ● | ● | | ● | ● | ● | ● | ● | ● | ● | ● | ● | | ● |
| 9 | Physical activity assistance | Assisting the elderly user to perform physical activities, such as walking | | | | | ● | | | | | | | | ● | | | | ● | | ● | ● |

### 4.1 CompanionAble (*Integrated Cognitive Assistive and Domotic Companion Robotic Systems for Ability and Security*)

The CompanionAble project [40] has linked smart home systems with "Hector," a robot that the project team claims to be fully autonomous. Hector is designed to play the role of companion for elderly people, especially those living alone or spending many hours of the day alone [20]. The objective is to help them remain independent, secure, fit, and happy through fall detection mechanisms integrated with emergency calls and remote monitoring services, and personalized dialogue/interaction displaying emotional intelligence (using visual, vocal, and tactile interfaces, sensor-based movements such as "follow me," and natural language recognition of commands) to mitigate feelings of loneliness. Hector provides friendly reminders, keeps and brings important objects, such as keys and wallets, offers cognitive stimulation and games, and provides seamless video connections to family and friends [20, 41–43].

### 4.2 DOMEO (*Domestic Robot for Elderly Assistance*)

To improve the wellbeing and autonomy of older adults, the DOMEO project [21] has developed mobility assistive and companion robots to provide personalized domestic services [44]. In this project, funded by the Ambient Assisted Living Joint Program of the European Union, the Kompai robotic platform has been successfully implemented in real homes with real people to accommodate those suffering from cognitive decline [45]. The DOMEO project aims to reduce long-term care costs. It does not replace human workers, but provides remote presence functionalities, such as monitoring and alarms, and helps care personnel and relatives better assist elderly individuals [46]. The objectives of the project are to provide cognitive and memory assistance, send a video stream in case of an emergency alarm, provide stimulation for physical exercise, and observe user behavior. The total project budget was 2.4 million euros [21].

### 4.3 Mobiserv (*Integrated Intelligent Home Environment for the Provision of Health, Nutrition, and Mobility Services to Older Adults*)

The EU-funded Mobiserv project [22] was organized to create a robot companion for older adults that can offer structure throughout the day, help people remain active by suggesting a variety of activities, and remind them to eat, drink, and take medicine. This venture had eight partners from seven countries, representing care organizations, universities, research institutions, and industry [47]. The objective of the Mobiserv project was to design, develop, and evaluate technology to support the independent living of older adults as long as possible, in their home or in various degrees of institutionalization, with a focus on health, nutrition, wellbeing, and safety [43, 48–51].

### 4.4 ExCITE (*Enabling Social Interaction Through Embodiment*)

The ExCITE project [23] methodology was inspired by a user-centric approach used for prototyping, validating, and refining a solution for elderly care via robotics. For the results of the evaluations to be significant, prototype deployment is considered on a large scale and from a longitudinal perspective [52–54]. Evaluation took place by utilizing a Giraff robot [55] prototype designed to accommodate future needs. Members of the ExCITE project were geographically distributed across Italy, Spain, and Sweden, and end-user participation is closely tied to the consortium and project activities. Healthy adult volunteers were selected at different end-user test sites. Each end-user site received a prototype to be tried and used for a period of up to one year, and the feedback shows very positive responses from the elderly and their families. Challenges in penetrating organizations have been outlined. The main objective of ExCITE was to evaluate user requirements for social interaction that enable embodiment through robotic telepresence. This evaluation was performed on a Pan-European scale with a longitudinal perspective [23].

### 4.5 ACCOMPANY (*Acceptable Robotics Companions for Aging Years*)

The ACCOMPANY system [24] includes a robotic companion as part of an intelligent environment, providing services to elderly users in a motivating and socially acceptable manner that facilitates independent living at home [56, 57]. The ACCOMPANY system provides physical, cognitive, and social assistance in everyday home tasks, and contributes to the reablement of the user, assisting them to carry out certain tasks on their own. This project is based on designing computational models for social cognition and interaction of robot companions. This process is aimed at delivering services through socially interactive, acceptable, and empathic interaction. The project plans to combines a multidisciplinary consortium to handle the technological, human-centered, and ethical challenges of the elderly care robotics.

### 4.6 HOBBIT (*The Mutual Care Robot*)

The HOBBIT project [25] envisions a robotic product that enables older people to feel safe and stay longer in their homes by using new technology, including smart environments (ambient assisted living; AAL). The main goal of the robot is to provide a "feeling of safety and being supported" while maintaining or increasing the user's feeling of self-efficacy (their own ability to complete tasks) [58]. Consequently, the functionalities focus on emergency detection (mobile vision and AAL), handling emergencies (calming dialogues, communication with relatives, etc.), and fall prevention measures (keeping floors clutter free, transporting small items, finding and bringing objects, and reminders). High usability, user acceptance, and a reasonable level of affordability are required to achieve the sustainable success of the robot [59, 60]. To achieve the goal of high user acceptance, the core element of the HOBBIT project is the concept of mutual care. Mutual care is an interaction design framework for assistive robots to facilitate relationships with their users [61, 62].

### 4.7 GiraffPlus

GiraffPlus [26] is a complex system that can monitor activities in the home using a network of sensors in and around the home, as well as on the body [63]. The sensors can, for example, measure blood pressure or detect whether somebody has fallen. Different services, depending on the individual's needs, can be preselected and tailored to the requirements of both older adults and healthcare professionals [64]. At the heart of the system is a unique telepresence robot, Giraff, which lends its name to the project. The robot uses a Skype-like interface to allow relatives or caregivers to virtually visit an elderly person in their home [65].

### 4.8 Robot-Era (*Implementation and Integration of Advanced Robotic Systems and Intelligent Environments in Real Scenarios for the Aging Population*)

The objective of the Robot-Era project [27] is to develop, implement, and demonstrate the general feasibility, scientific/technical effectiveness, and social/legal plausibility and acceptability by end users of a complete advanced robotic service integrated into intelligent environments. The service will actively work in real conditions and cooperate with real people to facilitate independent living, hopefully improving quality of life and efficiency of care for elderly people [66]. The Robot-Era project was designed based on an ambition to design, implement, and validate a set of robotic services for "aging well." It aims to tackle fundamental scientific and technological challenges in robotics and ambient intelligence to design cognitive-inspired robot learning architectures, taking into account elderly users' needs, design for acceptability, and legal/insurance regulations and standards for real deployment [67–69].

### 4.9 SILVER (*Supporting Independent Living for the Elderly Through Robotics*)

The SILVER project [28] searches for new technologies to assist elderly people in their everyday lives. By using robotics or other related technologies, the elderly can continue living independently at home, even if they have physical or cognitive disabilities. The first objective of the project is to establish and execute an agreed-upon pre-commercial procurement (PCP) process to run a cross-border PCP call for tender. This generic process should also form the basis for national PCP calls designed outside the SILVER project [70]. The goal is that in the future, public organizations in the participating countries and in the EU will be familiar with the PCP process and tools and use them to meet their needs. The second objective is to use the PCP process developed in the project to identify new technologies and services to support independent living for the elderly [71]. The project began by developing a generic PCP process and documentation, which will be used later as the basis for a specific call for independent living. The actual PCP process is executed in three phases. The first phase is a feasibility study of the selected technologies and proposals. The most promising ideas will be developed into well-defined prototypes in phase two, and the third phase aims to verify and compare the first real end products or services in real-life situations [72].

### 4.10 VictoryaHome

VictoryaHome [29] claims to monitor health and safety, facilitate social contact, and create peace of mind. Its services do not depend on automated functions, but empower family and friends, and bring immediate human presence when needed. It includes smart devices, such as an activity monitor, a fall detector, an automatic pill dispenser, a smartphone app for family and friends, an online dashboard for response centers, and a mobile telepresence device that stays with the senior [73]. The vision of this project reads, "Be Well – VictoryaHome – Create Possibilities." VictoryaHome is not only about responding to specific health or wellbeing problems, but also promotes self-care, being in control, and creating one's own solutions. This project provides the technology to connect elderly people to family and friends, to choose which support systems they would like to have, and to live their lives the way they want to. It is the project's mission to support people in taking care of each other and to bring peace of mind to all users. The VictoryaHome system and services have been developed in a user-centered design approach and tested and evaluated with a large group of end users in long-term trials in their own homes and social environments. Family, friends, and professional caregivers in four European countries have been included in the design, and a business strategy has been developed that is intended to fit within existing care models [74, 75].

### 4.11 TERESA

The TERESA project [30] aims to develop a telepresence robot with unprecedented social intelligence, thereby helping pave the way for the deployment of robots in settings such as homes, schools, and hospitals that require substantial human interaction [76, 77]. In telepresence systems, a human controller remotely interacts with people by guiding a remotely located robot, allowing the controller to be more physically present than with standard teleconferencing [78]. The project team has developed a new telepresence system that frees the controller from low-level decisions regarding navigation and body poses in social settings. Instead, TERESA has the social intelligence to perform these functions automatically. TERESA semi-autonomously navigates among groups, maintains face-to-face contact during conversations, and displays appropriate body-pose behavior [79].

### 4.12 RAMCIP (*Robotic Assistant for Mild Cognitive Impairment (MCI) Patients at Home*)

The RAMCIP project [31] aims to research and develop real solutions for assistive robotics for the elderly and those suffering from mild cognitive impairments and dementia. This is a key step in developing a wide range of assistive technologies. RAMCIP will adopt existing technologies from the robotics community and fuse them with user-centered design activities and practical validation, with the aim of creating a step-change in robotics for assisted living [80, 81]. The partners involved in the RAMCIP project range from experienced medical technologists to small- and medium-sized enterprises (SMEs) developing cutting-edge robotics. They have come together with the common goal of enabling a new generation of assistive technology, drawing on both their extensive experience and the enabling effects of European support. The work of the project is specific, focused, deliverable, and will supposedly advance state-of-the-art technology in a measurable way. The partners have identified routes to market for the project output, along with economic models that offer great confidence that a system of real, concrete impact on one of Europe's key societal challenges can be created [82].

### 4.13 ENRICHME (*Enabling Robot and Assisted Living Environment for Independent Care and Health Monitoring of the Elderly*)

The acronym of the project "ENRICHME" [32] comes from "ENabling Robot and assisted living environment for Independent Care and Health Monitoring of the Elderly." It refers to the goal of enriching the day-to-day experiences of elderly people at home by means of technologies that enable health monitoring, complementary care, and social support, helping them remain active and independent for longer and enhancing their quality of life [83–85]. ENRICHME is not intended as a substitute for human contact or social activities, but as an enabling tool for social interactions and inclusion, utilizing the capacity of the robot and the wider connectivity that it allows.

### 4.14 GrowMeUp

In the GrowMeUp project [33], the main aim is to increase the years of independent and active living and the quality of life of older people with light physical or mental health problems (from the age of 65+) who live alone at home and can find pleasure and relief in getting support or stimulation to carry out their daily activities [86, 87]. GrowMeUp aims to provide an affordable robotic system that is able to learn an older person's needs and habits over time and enhance (or "grow up") its own functionality. In doing this, it compensates for the user's deteriorating capabilities to support, encourage, and engage them to stay active, independent, and socially involved while carrying out their daily living activities at home. State-of-the-art cloud computing technologies and machine learning mechanisms will be used, enabling the GrowMeUp service robot to extend and increase its knowledge continuously over time, as well as potentially sharing and distributing its knowledge with multiple other robots through the cloud. In this way, robots making use of the cloud can learn from each other's experiences, thus increasing their functionality and competencies while reducing learning effort [88, 89].

### 4.15 Mario (*Managing Active and Healthy Aging with Use of Caring Service Robots*)

The Mario project [34] addresses the challenges of loneliness, isolation, and dementia in older people through innovative and multi-faceted inventions delivered by service robots. Human intervention is costly, but this can be prevented and/or mitigated by simple changes in self-perception and brain stimulation mediated by robots [90]. Considering the role of a robotic system, clear advances could emerge in the use of semantic data analytics, personal interaction, and unique applications tailored to better connect older people to their care providers, communities, social circles, and personal interests [91]. Each objective is developed with a focus on loneliness, isolation, and dementia [92].

### 4.16 RADIO (*Robots in Assisted Living Environments*)

The RADIO project [35] pursues a novel approach to acceptance and unobtrusiveness: a system in which sensing equipment is not discrete, but an obvious and accepted part of the user's daily life. By using the integrated smart home/assistant robotic system as sensing equipment for health monitoring, users' attention is redirected to the functionality of the sensors rather than to the sensors themselves [93]. In this manner, sensors do not need to be discrete and distant or masked and cumbersome to install; they do, however, need to be perceived as a natural component of smart home/assistant robot functionalities [94]. The project is centered on the RADIO Home, an integrated smart home and assistant robot clinical monitoring environment in which the hardware primarily serves user comfort and home automation purposes [95–97].

### 4.17 SocialRobot

The main goal of the SocialRobot project [36] is to provide a solution to the demographic change challenge through knowledge transfer and the creation of strategic synergies between the project's participating academia and industry, with the goal of creating an integrated social robotics system (SocialRobot) for "Aging Well." The work focuses on bringing together the robotics and computer science fields, integrating state-of-the-art robotic and virtual social care communities' technologies and services to provide solutions to key issues of relevance for improved independent living and quality of life for elderly people. The SocialRobot development will be based on a "human-centered approach" in which the elderly individual's needs are met. The project will provide an opportunity for participating SMEs with excellent credit in their domains and peripheral European regions to achieve excellence and compete with innovative products in the elderly care market at the European and international levels [98]. The major challenges to be addressed in the project include the adaptation of state-of-the-art robotic mobile platforms and their integration with a virtual collaborative social network to provide the following: detection of individual needs and requirements related to aging (e.g., physical mobility limitations and cognitive decline); provision of support through timely involvement of care teams, consisting of different groups of people (family members, neighbors, friends) that collaborate dynamically and virtually regardless of time and physical location; behavior analysis to adapt to the social relationships of elderly people as they age; and navigation of indoors and unstructured environments to provide affective and empathetic user–robotic interaction, taking into account the capabilities of, and acceptance by, elderly users [99, 100].

### 4.18 MECS (*Multimodal Elderly Care System*)

The main goal of the MECS project is to investigate the development of technology for a safety alarm robot that detects non-normal situations and thereby allows for the independent living of older people [37]. No specific robotic platform was available at the start of the project to investigate the use of safety alarm robots in the homes of the elderly. Thus, the project first explored what a robot meant for the elderly and how they understood it [101]. During this first phase, it was observed that the elderly were reluctant to use robots that were large in size and those that would surveil them. In the next phase of the project, Sony Aibo, NAO, and TurtleBot2 were tested with elderly people [102]. Sony Aibo and NAO demonstrated some companionship elements. The elderly users were positive about these robots; however, they did not see any real use for them apart from having them as companions. TurtleBot2 was used in an experiment with elderly participants [103], and it was found that users preferred to use finished, off-the-shelf products rather than unfinished research platforms. They also preferred servant and assistive robots rather than “care” or “safety” alarm robots that monitored their health state and progress. A number of elderly participants were willing to use off-the-shelf vacuum cleaner robots in a longer-term user study. MECS provided each of the participants with one of three different robotic vacuum cleaners after testing them in pilot studies. Neato and Roomba were distributed among the elderly and used for a period of one month. Non-elderly participants were also asked to use the robots and document their experiences. The documentation included diary notes, logs, and photos, which were later used in interviews with the participants.

Another exploration of this project involved the Fetch robot [104], for which the animation principle “slow-in-slow-out” was tested in a living lab at the University of Hertfordshire [105, 106]. Users’ feedback about interacting with a Pepper robot was also registered and analyzed at the VITALIS care facility in Eindhoven and compared to young users’ perceptions of the robot [107].

### 4.19 VIROS (*Vulnerability in the Robot Society*)

The VIROS project explores the legal and regulatory questions introduced by the deployment of robots and artificial intelligence (AI) systems in healthcare contexts [38]. VIROS aims to regulate robotics from a technical and legal point of view by addressing the safety, privacy, and security aspects that these systems raise and their deployment in private and public spaces. Researchers in robot engineering, law, and social sciences, in conjunction with designers, examine how robots can be integrated into less controlled environments, such as homes or hospitals. The project addresses specific legal issues, AI regulation and risk assessment, regulatory aspects with respect to the General Data Protection Regulation (GDPR), informed consent and data minimization, design elements such as universal design and inclusion, and robot engineering aspects, such as algorithms for improving the performance of robot navigation and control in human–robot cooperative tasks. The project aims to test humanoid robots, such as EVE, Lio, and TIAGo, in natural settings, such as hospitals with nurses and patients and the homes of the elderly.

### 4.20 HIRO (*Human Interactive Robotics for Healthcare*)

HIRo [39] introduces a new concept for improving health services in hospitals and municipalities through the integration of a humanoid robot (EVE). The goal is to develop a robot that can act as an assistant to personnel and patients in daily tasks, such as transporting or handing over equipment for medical tasks, helping patients find their way in facilities, or assisting them to pick up objects or move around. The interdisciplinary team was put together specifically for this project and includes health personnel in hospitals and municipalities, pioneering robot developers, and researchers with knowledge of patient rehabilitation, interaction, and safety between people and machines. In this project, the team works closely with health professionals and patients in hospitals and in the municipality to identify the needs, desires, and expectations for interaction with a humanoid robot.

## 5 FINDINGS

This paper has analyzed 20 research projects focusing on the application of robotic systems in elderly care. This section provides an overview of the common themes identified among the projects: A) the care robot’s functionalities; B) social- or tele-robots as products for different care receivers in non-controlled settings; and C) design and ethical aspects of care robots. We present each of these below.

### *5.1* The Care Robot’s Functionalities

The majority of the research projects (n = 17, ID = 1, 2, 3, 4, 6, 7, 8, 10, 11, 12, 13, 14, 15, 16, 18, 19, and 20, shown in Table 1) focus on the robot’s functionalities and roles, as well as its technical equipment. In these projects, the robot’s functionalities were often seen as important and relevant for physical and cognitive stimulation to keep the elderly active, increase their quality of life, and address their needs. The same robots were also seen as compensating for the elderly’s declining abilities. Specifically, the projects’ main focus was on what the robot should do and how it should behave, for example, whether it should bring objects, act as a communication channel for conversations with family members, detect or prevent falls, or make emergency calls. Other functionalities and roles of the robot concerned whether it should be able to provide a personalized dialogue or possess emotional intelligence.

These projects focused on multimodal types of interaction that the robot should have, such as providing reminders, alerts, or alarms; monitoring the user’s state and/or acting as a social companion; motivating the user to do some exercise; reading their facial expressions and emotions; or acting as part of a larger, more complex system, such as a smart home. Other functionalities included the robot providing cognitive assistance, recording patterns in human behavior (including sleep patterns), detecting obstacles, and providing entertainment, such as updating human users with information about the latest movies, books, and music, or providing access to social media. There was also a focus on the robotic system providing advanced indoor and outdoor tasks safely. In some cases, the robot was designed to mimic the human user, such as by nodding the head, while other projects focused on the dexterity of the robotic arm and its ability to grip objects, moving them from one place and to another safely.

With regard to technical equipment, the projects often talked about what kinds of sensors or extra equipment the robots should include. Significant focus was given to extra equipment, such as different types of cameras, speakers, and microphones, along with whether the robot should have an embedded screen or a touch screen, medical or non-medical sensors (e.g., oximeters or physiological sensors), occupancy sensors or other universal sensors, or different plugins, such as augmented reality plugin sensors or computer vision sensors.

Another important consideration encountered in the analyzed research projects is related to indoor navigation in non-controlled settings, such as private homes. Robots often encountered static and dynamic obstacles that made it difficult to navigate the environment. An additional technical challenge discussed by some of the projects (ID = 4, 7, 17, and 18) was related to how robots could navigate a physical environment safely by following paths through doorways and narrow passages, regardless of whether they were tele-operated or not. The issues of autonomous navigation and the robot's ability to dock itself were often problematic.

Some of the research projects (ID = 7, 10, 11, 18, and 4) focused on how they could control the robots remotely by having effective and efficient control architecture that allowed them to be easily maneuvered. In these cases, the controller was human, and the robot was expected to respond within seconds.

More than half of the projects (ID = 4, 5, 6, 7, 8, 9, 10, 12, 15, 16, 17, and 20) referred to and collected information about user requirements. Among the requirements listed were the adjustable height of the robot, pre-selection of services, social and legal acceptability of the robot, legal insurance of the human user, the need for regulations and standards, the need for the robot to cooperate and operate in domestic settings, robots' use in domestic and urban contexts in everyday life, and the need for the robot to understand the behavior of multiple people in the user's home. Other requirements covered the compatibility of the robot with the rest of the technical system, including how it should be integrated within a larger system. Many of the projects claimed to address the users' individual needs, and some focused on the actual process of understanding these needs in order to map them.

In terms of technical solutions, machine learning techniques and semantic analysis methods were used (ID = 15, 16, 18, and 19). A summary of the different functionality aspects of the studied projects is shown in Figure 4.

### *5.2* Robots as Products in Care Settings/Robot as an isolated care element

Many of the projects (n = 16, ID = 1, 2, 4, 5,6,7, 8, 10, 11, 12, 13, 14, 15, 16, 17, and 20) focused on the concept of care, such as elderly care, independent living, aging well, or healthy aging. Aging was a key demographic component of the projects. The focus on care was indicated as elderly ICT-supported care, therapy management, multimodal user observation, telehealth and virtual care, integrated care scenarios, homecare, networked care, personalized homecare services, and cognitive and physical assistance for the elderly. The majority of the projects aimed to delay elderly people's moves into care homes and intensive care by increasing quality of life and efficiency of care. Integrating robotic systems as part of elderly care was seen as a new healthcare paradigm that aimed to avoid long-term institutional care, instead providing home and remote social care services.

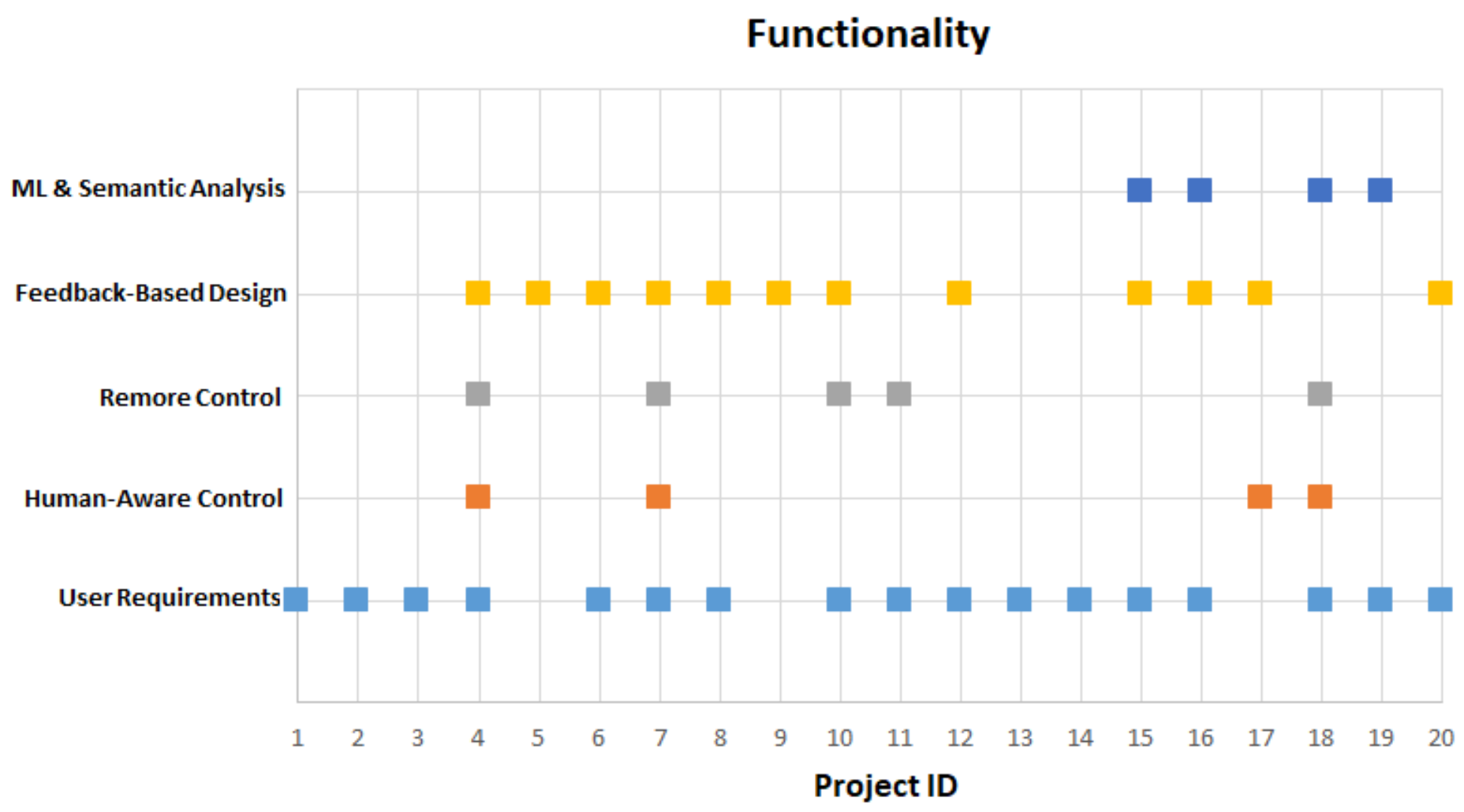

Figure 4. Summary of the functionality aspects of the projects.

The users in the projects were the elderly, as well as formal and informal caregivers, people with mild cognitive and physical disabilities, mild cognitive impairments patients, healthcare professionals, patients, and different stakeholders. However, only three of the projects (ID = 5, 6, and 10) addressed human autonomy in terms of user empowerment, and how the robot could physically, cognitively, and socially empower the human user with a user-centered approach.

In some of the projects, the deployment of the robot was aimed to be in real-world contexts, including dynamic and cluttered homes. In these cases, the robot must be able to operate even in low lighting conditions, with users with different skin colors, and in environments with different types of carpets and floors. In other projects, the deployment of future robots was for settings such as schools, hospitals, elderly housing facilities, labs, or other assistive living environments. Eight of the projects discussed robots as an integrated part of smart home environments, or as part of an automated and intelligent system. However, the testing of such robotic systems was mainly conducted in realistic experimental setups, field experiments, or in test homes, where robots were tested over shorter or longer periods of time, rather than adopting a holistic approach in which the robots were part of an integrated health or home care service. Figure 5 shows different aspects of the robots' nature in these projects.

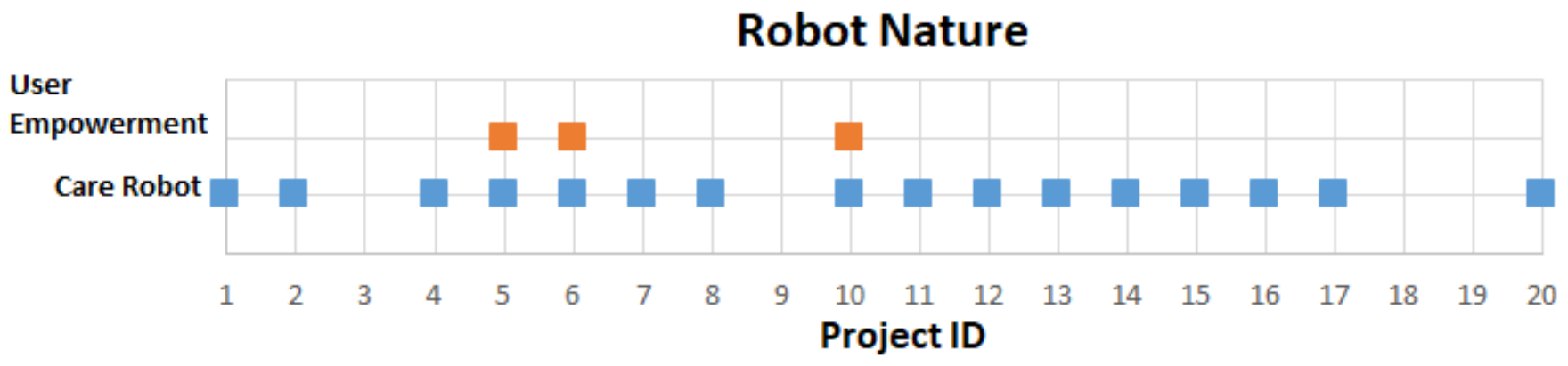

Figure 5. Summary of the functionality aspects of the projects.

### 5.3 Strong Emphasis on Robots as Commercial "Products"

Almost all projects (n = 17, ID = 1, 2, 4, 5, 6, 7, 8, 9, 10, 12, 13, 14, 15, 16, 17, 19, and 20) considered robotic systems as future products to be part of elderly care, with either an innovative or commercial approach in mind. The robotic systems were presented as novel technological solutions that underwent several prototyping versions, and that aimed to reach the elderly care market or were designed in such a way that they were close to commercialization. Other projects focused on the process of "the way to the market" rather than the robot product itself, and aimed to develop PCP processes, tools, business strategies, and business knowledge around different possible purchasers at different levels. Some of these projects also focused on feasibility studies and conducted testing in controlled settings to make a risk assessment of the robot's viability. One project even included 33 tenders (ID = 9), while another aimed for 10% penetration in care organizations (ID = 10). At the same time, other projects focused on SMEs and the creation of cutting-edge robot systems designed as integrated platforms and their routes to market, and economic exploitation in terms of technology for healthy aging. Some projects aimed to harness the potential of collaboration to develop products for use in the European Union scientific and commercial market for service robots while looking for competitive advantages for developing commercial footprints, which would eventually develop into a wide deployment of technology in the elderly care market. The end users were seen as consumers.

Half of the projects (n = 10, ID = 1, 4, 5, 6, 10, 12, 13, 14, 15, and 16) had economic drivers behind them. These projects were driven by the economic pressure of expensive robot products for the elderly, and aimed to create affordable products for elderly people with limited financial resources. Some of these projects focused specifically on decreasing the costs of elderly care, noting that human intervention was costly. These also focused on the creation of economic models aimed at lowering elderly care costs, and on cost-effective solutions for the creation of robot products to be integrated into care.

Four projects (ID = 5, 9, 12, and 13) focused on the process rather than on the robot product as such. The focus was either on evaluating robot products or on different development cycles that a robot should go through before reaching the market. Only one project (ID = 4) focused on assessing the robustness and validity of robots using short- and long-term longitudinal studies on multiple test sides, and only one project aimed to validate its results in real-life scenarios rather than through experiments (ID = 12). Moreover, only one project focused on the impact of integrating a robot as part of elderly care (ID = 13), rather than on the robot as a technical product, and only one focused on sustainability aspects of the robot, such as its battery life (ID = 4). The same project (ID = 4) and one other (ID = 6) focused on the usability aspects of the robot and user acceptance, and eventual problems with its utility. This project (ID = 4) performed usability tests, and concluded that the participants using the robot needed time to become familiar with the device.

All the projects (N = 20) focused on some type of robot, from companion robots, such as Hector, to robot assistants or mobile robot companions (e.g., ID = 1). Other projects focused on off-the-shelf robots and open robot platforms, such as Kompai (e.g., ID = 2 and 15). Robot companions were quite popular among the projects, including Giraff (e.g., ID = 4, 7, and 11), which is a domestic mobile robot telepresence system. In these projects, the focus was on developing the technical system itself through a number of prototypes. Other projects focused on service robots that were socially intelligent systems, socially assistive robots and futuristic humanoid robots like EVE (ID = 20), or non-humanoid robots with high levels of cognition. Only one project considered connected robots and how several robots could share and distribute their knowledge through the cloud (ID = 14).

Around half of the projects (n = 11, ID = 1, 3, 4, 7, 10, 11, 13, 14, 15, 16, and 17) took account of the risk of social exclusion or isolation and how robots could mitigate that risk. Social interaction, social cognition, social connectedness, and remote interaction were at the heart of many projects. These projects focused on social- or tele-robots that aimed to provide the elderly with a way of being social, either by helping them keep in touch with family and other caregivers, or by direct interaction with a robot companion. The human and the robot aimed to coexist in the same space, where the robot provided an empathetic communication channel, having an emotional understanding and being able to communicate in an intelligent way in order to mitigate feelings of loneliness and social isolation. One concern addressed in these projects was the risk of social or companion robots replacing people, thus leading to the elderly losing even more social contact with other humans. Figure 6 summarizes the different concerns discussed in consideration of the care robot as a product.

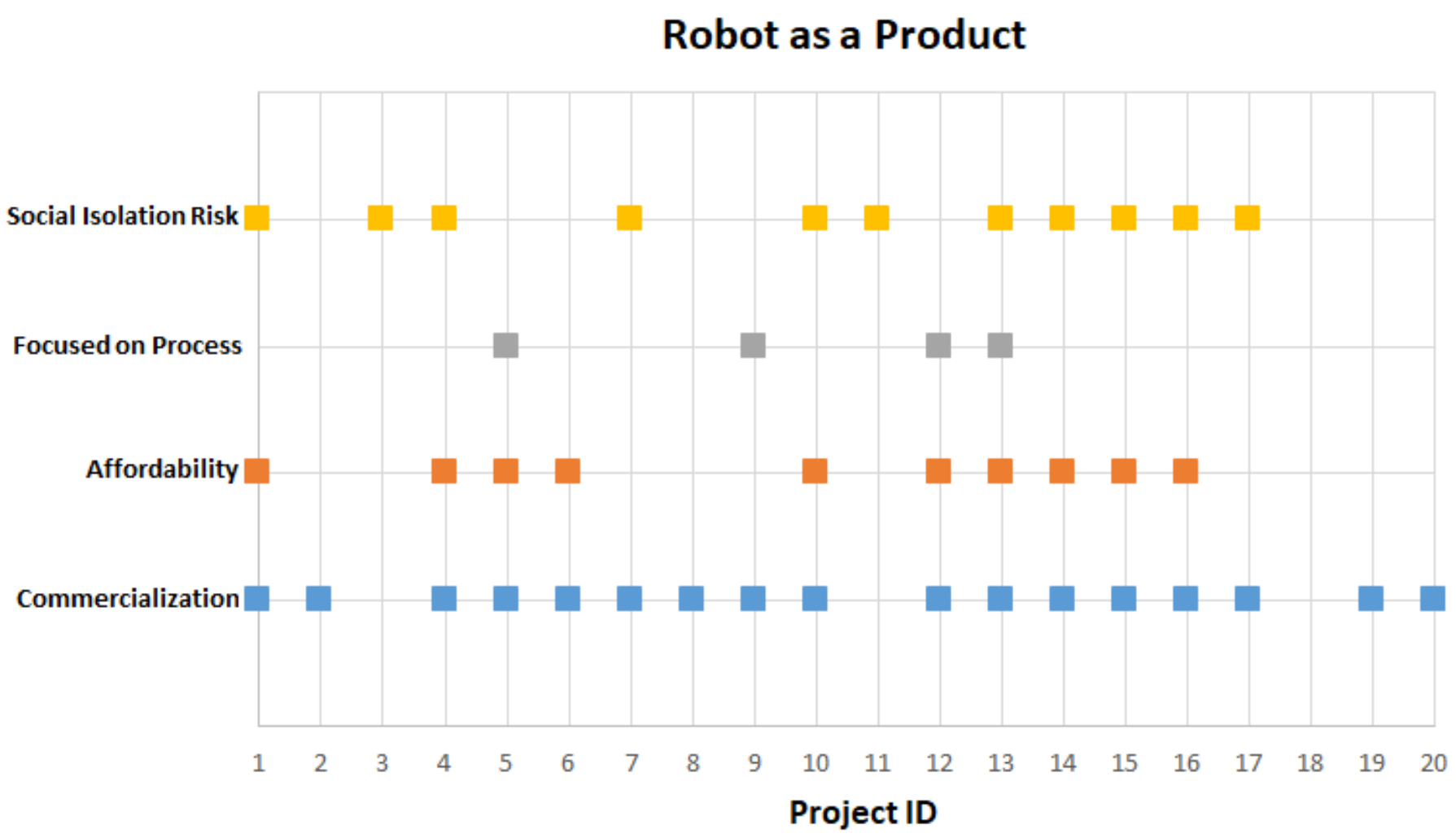

Figure 6. Summary of findings related to robots as a commercial product.

### 5.4 Design and Ethical Aspects of Care Robots

Less than half of the projects (n = 9, ID = 2, 3, 4, 5, 7, 14, 15, 18, and 19) had extensive regard to ethical aspects, such as issues related to data collection, data sensitivity, and related privacy concerns, or ethical guidelines, challenges, and frameworks. In some of these projects, discussions on ethics were mainly concerned with data and how it was stored. Only one project (ID = 19) focused specifically on legal issues with regard to digital vulnerabilities, safety, privacy, and security aspects, alongside its focus on technical aspects.

Most projects focused on the elderly's state and how a robot could keep them physically and mentally fit, happy, active, and independent. The elderly were seen as users with different medical conditions, often perceived as depressed, tired, suffering from dementia, at risk of falling, and undergoing cognitive decline. They were also seen as having mobility problems, deteriorating health conditions, and chronic disabilities. The robots in these contexts aimed to mitigate these challenges and offer a solution by providing reablement, facilitating independent living at home, enhancing wellness, and improving quality of life. The robot and the user were supposed to mutually assist and take care of each other, with the user regarding the robot as a partner. In terms of safety, a few projects (n = 8, ID = 2, 4, 6, 7, 12, 18, 19, and 20) addressed safety issues concerned with the physical safety of the user when the robot was navigating and moving around the home environment or undertaking dexterous arm manipulation during handover tasks. None of the projects focused on psychosocial aspects of safety alongside physical aspects.

One (perhaps contested) focus area of some of the projects (n = 8, ID = 2, 3, 7, 10, 12, 15, 16, and 18) was the idea of "surveillance," that is, long-term, systematic monitoring of human health and behavior, with the robot not only knowing the state of the human user, but also understanding the home environment. Only three projects (n = 3, ID = 5, 6, and 10) focused explicitly on empowering human autonomy and having a human-centered or user-centered approach, in which the human would be able to make their own choices with regard to the use of the robotic system and their own health and wellbeing. Rather, attention was on the robot's autonomy and technical solutions for making the robot more autonomous. Only one project (ID = 6) focused on the robot's accessibility.

Few projects gave much explicit consideration to the feelings of the user, such as possible feelings of frustration or attachment to the robot (which could be similar to an attachment to pets), and the risks of developing real feelings of affection toward the robot. Only one project (ID = 4) focused on the user experience of feeling safe around the robot. Figure 7 shows four main considerations for dealing with the ethical aspects of designing the care robot, based on the studied projects.

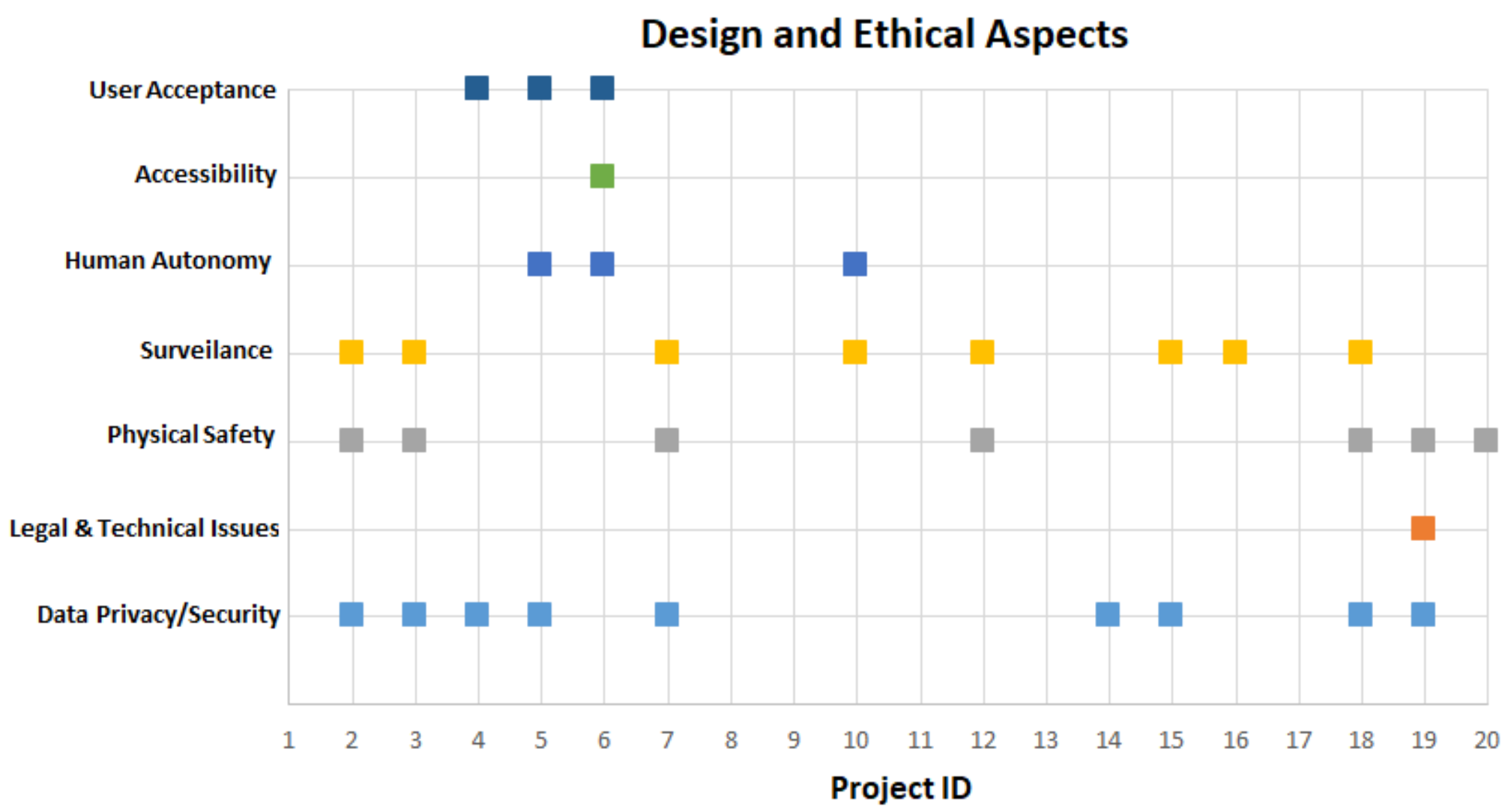

Figure 7. Summary of the design and ethical aspects of the projects

Service robots have been used in factories and laboratories for some time and are "good at dull, dangerous, and dirty work," such as cleaning, doing domestic work, harvesting fruit, assisting surgeons, and performing activities that are dangerous for humans due to chemical exposure [108]. Caregiving is not, however, like factory or laboratory work. It is a complex and privileged process that requires a high level of knowledge and skill. It also requires a high level of ethical sensitivity and a commitment to preserving the dignity of care recipients in every context. The risk of disrespect and alienation in such a context is very real. Older people are particularly vulnerable to dignity loss if they become dependent and lose some capacity for decision making. They are reliant on caregivers to respond to their individuality, pick up on subtle cues regarding capabilities, and communicate in an emotionally engaged and meaningful way. This involves being present to the person receiving care, making them comfortable, and enabling them to feel that they are valuable, regardless of any incapacity. Based on our analysis, we compiled a list of the main ethical issues addressed across projects. An extensive study on ethical issues for care robots is provided in [109]. Table 4 summarizes our findings on these ethical issues.

Table 4. List of main ethical issues

| # | Issue | Description |
|---|---|---|
| 1 | Potential reduction in the amount of human contact | Opportunities for human social contact could be reduced, and elderly people could be more neglected by society and their families than before. Robots could provide an excuse for such neglect if others mistakenly choose to believe that the seniors' physical and emotional needs are being taken care of by machines. |
| 2 | Increased feelings of objectification and loss of control | The insensitive use of robots developed for the convenience of carers could increase elderly people's feelings of objectification and loss of control. This could occur if, for instance, robots were used to lift or move people around without consulting them. |
| 3 | Loss of privacy | Loss of privacy and a restriction of personal liberty could result from the use of robots in elderly care. The extent to which robots should be allowed to restrict the behaviors of humans is a subjective question, with implications beyond care of the elderly. |
| 4 | Loss of personal liberty | |
| 5 | Increased feelings of alienation | The risk of reduced human social contact may lead to the elderly experiencing feelings of alienation. |
| 6 | Deception and infantilization | Deception and infantilization of elderly people might result from encouraging them to interact with robots as if they were companions (as discussed, the possible willing collusion of the elderly with such deception makes this issue a complex one). |
| 7 | Circumstances in which elderly people should be allowed to control the robots | If robots are placed under the control of elderly people, there is an issue of responsibility if things go wrong. This opens up other important issues, such as the extent to which the wishes of the elderly person should be followed, and the relationship between the amount of control given to the elderly person and their state of mind. |

## 6 DISCUSSION

This paper has analyzed 20 research projects on the use of robotics in elderly care. This section discusses: A) the current progress and commonalities among the projects; B) the main challenges and opportunities presented by the use of robots and robotic systems in elderly care; and C) future directions with regard to the shortcomings of the current projects, and what similar projects should address next. We discuss each of these below.

### 6.1 Current Progress and Commonalities Among Robots in Elderly Care

The central findings of our analysis can be summarized as: (i) the majority of the projects focused on care robots' functionalities and on the technical equipment (n = 17); (ii) robots (social- or tele-robots) were seen as products in care settings for different care receivers; (iii) a strong emphasis was placed on robots as commercial products; and (iv) less than half of the projects (n = 9) focused on the ethical aspects of robots. In the paragraphs that follow, we discuss each of these central findings.

First, most of the projects had central regard to the technical side, such as the robots' functionalities and the technical requirements for a viable product. In other words, the focus was on which functionalities a robot system should include in order to be commercialized. Thus, the robot systems for elderly care were mainly seen as products with potential for economic gain, rather than an element of care. Similarly, the majority of the projects focused on processes, such as "the way to market," and the provision of personalized robots, emphasizing this product-oriented focus and commercialization. Missing from these projects was again the element of care as a central focus. The concept of care is understood from a relational perspective, in which at least a caregiver and a care recipient are involved, but it is also a moral and political concept [110] with practical dimensions. How we talk about care robots at a conceptual level and how we integrate them into our care processes have consequences at the social and ethical levels. For instance, [111] defines care robots as "robots intended to assist or replace human caregivers in the practice of caring for vulnerable persons such as the elderly, young, sick, or disabled" (p. 251). Robots cannot be considered caregivers or care providers in a phenomenological way of understanding the concept. Robots can, at best, support care processes as elements of care. Similarly, [112] emphasizes the problem of current policies primarily addressing the economic impact of robotics, focusing on physical safety, for instance, while disregarding social and cultural implications and what it actually means to interact with a robot over a long period of time. In addition, [113, 114] address the issue of responsible robotics, shifting the focus from potential economic gain toward which tasks should be allocated to robots and which should continue to be performed by humans. This is an important paradigm shift in our understanding of care robots because it changes our fundamental understanding of how they should be designed and integrated in elderly care.

Second, there was a focus on social- or tele-robots as products for different care recipients in non-controlled settings. In most of the projects, the robots were seen as either companions, used in homes or other non-controlled settings, or as supporting independent living through functionalities designed for mental fitness, keeping the elderly active, mitigating feelings of loneliness with personalized dialogue, or a combination of these. However, the actual meaning of a social robot as concept was taken for granted. In this regard, it is worth noting claims by [115] that research projects with social robots are often carried out by engineers who have limited knowledge about social aspects, including what sociality actually entails. The authors argue that robots cannot be social in the true sense of the concept, with its cultural and emotional dimensions. Robots can, at best, be sociomorphed; that is, they may appear to be social, but they are not capable of social and cultural interactions in the same way that humans are [116, 117]. Researchers often discuss the topic of asymmetric social interactions between humans and robots because of their incongruency in social capabilities [118, 119].

Finally, while some design and ethical aspects of care robots were touched upon by the projects, this occurred in a rather limited way. Legal aspects were largely missing from their respective agendas, which is unfortunate, as the legal regulatory framework sets important parameters for robotics development. This framework is also changing in ways that may significantly impact future work in the field. We see this, for example, with the European Commission's current proposals for an Artificial Intelligence Act (AIA) [120] and the Regulation on Machinery Products (RMP) [121]. Both of these, if enacted, will place new demands on research projects that build on AI-related applications in robot systems for elderly care. Thus, future projects should address legal factors and the practical challenges these create for robotics.

Much the same can be said with respect to the sustainability aspects of robotics, which were touched upon by some projects to a very limited degree (e.g., referring only to improvement in battery energy consumption). At a global level, the United Nations emphasizes the importance of sustainability, particularly with respect to its Sustainable Development Goals (SDGs) [122]. Several

SDGs are relevant for the design, development, and integration of robots in care processes, such as SDG3 on good health and wellbeing, SDG9 on industry, innovation, and infrastructure, SDG10 on reduced inequalities, SDG12 on responsible consumption and production, and SDG17 on partnerships for the goals.

### 6.2 Challenges and Opportunities of Healthcare Robots for the Elderly

One concern that must be highlighted is that while focusing on the benefits of introducing robots to elderly care, ethical concerns that might be introduced should not be neglected. Robots integrated into care processes should be seen as a part of large and complex sociotechnical systems with their own established practices. Robot systems should therefore not be considered in isolation, solely as products with potential economic gains, or as (the only) technical solutions to care-related challenges. As previously mentioned, they should be seen as elements of care in a large and complex social, economic, and cultural context.

Elderly people or other users, such as formal and informal caregivers, who will be utilizing these care robots should be seen as the main stakeholders in a user-centered design (UCD) or participatory design (PD) process, rather than part of isolated experiments. Some studies on robots used by the elderly show that the concept of robotics may be unfamiliar to the elderly, and that they often are pushed toward using modern advanced technologies without understanding them, leaving them feeling dismissed or bypassed [101]. Therefore, we suggest that the human–robot interaction (HRI) field investigating care robots for elderly users would benefit from design processes that involve users at all stages of the conception, development, and utilization of the systems concerned in the way that PD methodology does. PD has been extensively employed to study participation in design [123, 124]. Among PD values are democracy, mutual learning, and empowerment. Through such methodology, the elderly can have a greater voice in the design, implementation, adoption, and use process, rather than being consulted only in relation to the robot's functionalities.

Another challenge highlighted is that most projects did not focus on how AI could impact the privacy, safety, and security aspects of care robots. As indicated in the previous sections, this should be high on the agenda, particularly in light of the aforementioned legislative proposals for the AIA and the RMP.

Other challenges that were not addressed by the analyzed projects concern human rights and the need to create inclusive care robots that can be used by a diverse group of elderly people. Article 25 of the Universal Declaration of Human Rights [125] states: "Everyone has the right to a standard of living adequate for the health and wellbeing of himself and of his family, including food, clothing, housing, and medical care and necessary social services, and the right to security in the event of unemployment, sickness, disability, widowhood, old age or other lack of livelihood in circumstances beyond his control." Here, we are interested in the right to a standard of living that is appropriate for the health and wellbeing of individuals and their families, particularly in the case of aging or disability. This is important because it connects directly to the right to health [126], which encompasses "the right to a system of health protection providing equality of opportunity for everyone to enjoy the highest attainable level of health; Equal and timely access to basic health services; Participation of the population in health-related decision making at the national and community levels."

Discrimination must also be avoided, and all services should be made available and accessible to diverse groups, including the elderly, people with disabilities, and other vulnerable people. Accordingly, future research projects on care robots need to address the elderly as a diverse group. They must consider the risk of further marginalization if users have low digital literacy, rendering them unable to use or comprehend robot applications. Universal design (UD) [127, 128] principles of equitable use, flexibility in use, simple and intuitive use, perceptible information, tolerance for error, low physical effort, and size and space for approach and use should be addressed at both the physical and virtual levels. These principles have been shown to be useful in the design of physical environments, as well as in web and mobile applications. However, they seem to be underexplored within HRI and robotics, apart from the existence of a few available studies [129–131] and a recent call out from [132] regarding the need for these principles in robotics.

Value sensitive design (VSD) [133, 134] or care-centered value sensitive design (CCVSD) [135, 136] should receive more attention in future research projects on care robots for the elderly. Particular values need to be protected and embedded in technology development processes. One such value is care, as discussed at the beginning of this section.

Other concepts need to be readjusted because their meanings change with the adoption of future care robots. Concepts such as autonomy can evolve over time. For instance, nowadays, autonomy is sometimes understood merely as human vs. machine autonomy. Studies have shed light on the importance of understanding how human autonomy changes with the adoption of robots in societies [137]. Others highlight autonomy as a core concept [138] for enhancing human agency without devaluing human abilities, but also as a core principle for AI in society, regardless of whether the AI system has a physical layer. Another concept addressed in the literature is safety [139]. So far, the majority of research projects investigating robots have looked at safety from a physical perspective. However, for long-term interactions with robots, the cognitive and psychosocial dimensions of safety should also be addressed. In this way, some concepts may be subject to new valences. Other concepts that are relevant for investigation are human dignity and empowerment.

### 6.3 Technical Challenges and Opportunities

Our findings show that the research projects focused on the limitations and weaknesses of the care robots from a technical point of view, and how these could be improved with new sets of cameras, sensors, speakers, microphones, embedded displays, or plugins. The projects also explored aspects of navigation in static or dynamic settings and issues related to remote control. Some of the technical challenges and opportunities of these research projects addressed the use of reference architectures for social robots in order to utilize standardized software (or hardware) components for a series of robots, which may in turn foster interoperability among them [140]. Finally, robot-robot interactions, along with human-robot interactions, in which the human is part of a system involving several integrated (care) robots, would also be interesting to investigate.

### 6.4 Discussion of Ethical Aspects

With elderly care as a rapidly approaching problem worldwide, particularly in Europe and developed parts of Asia, most major robotics research institutes and universities are engaging in the home healthcare space. For example, the Fraunhofer Institute is working on the Care-O-Bot [141], and robotics associations in Japan [142], China [143] and Korea [144] have their own elderly care robot projects.

Current AI implementation in elderly care robots, however, is still in its early stages, and before robots completely substitute their human counterparts in care tasks, many technological and ethical hurdles need to be overcome. A report by UNESCO's World Commission on the Ethics of Scientific Knowledge and Discovery (COMEST) [145] states that the preservation of human dignity and privacy currently falls under ethically uncharted territory for robots. Some examples of ethically unclear situations that call for constraint or caution regarding robot deployment are described in the following paragraphs.

If an elderly care robot is tasked with reminding patients to take their medicine, the underlying robot intelligence needs to know what to do if the patient refuses to take them. This is especially difficult for current AI platforms since a patient may be refusing the medicines for a legitimate reason, or a robot that gives passive reminders to an old person might be considered impractical because they cannot replace a human nurse counterpart. Another example would be an elderly person having high-calorie foods taken away from them due to a robot-made decision to prevent obesity. Finally, situations could arise in which a caregiver uses a remote-controlled robot to restrain an elderly person, leading to moral and legal ambiguities.

## 7 CONCLUSION AND FUTURE WORK

This paper has analyzed 20 recent research projects on robotic systems for elderly care. We used the thematic analysis method [19] to analyze the dimensions addressed by the research projects. Our findings were discussed in terms of care robot's functionalities, social- or tele-robots as products for different care receivers in non-controlled settings, and design and ethical aspects of care robots. We have arrived at five key points that also act as a roadmap for future research projects addressing robotic systems for elderly people, which are summarized in this section.

Extensive work in the area of social and assistive robotic systems for elderly people has been undertaken. Significant developments have taken place in this field with regard to functionalities, technical development, and old and novel equipment being used to create advanced and intelligent robots. Considerable effort has been invested in the exploration of robots' different roles and behaviors. With social and assistive robots becoming even more relevant in our aging society, it is important that this work continue.

Although we recognize these great advancements, our study shows that the field of robotic systems for elderly people has room for improvement. To achieve greater deployment and use of robots in homes and healthcare for elderly people, and to meet the increasing demands of this population, we need to address other aspects of social and assistive robots.

1) We suggest that research projects on care robots for the elderly should shift their attention from being predominantly product-oriented and concerned with economic gain to focusing on the concept of care. These kinds of robots need to continue being developed and thought of as part of an already existing (home or health) care infrastructure. At the same time, the wellbeing of the elderly should take center stage rather than economic gain.

2) Future research projects should ensure greater involvement of the elderly as key stakeholders in the process of planning, designing, implementing, testing, adopting, and using care robots. Participatory design and other similar methodologies would facilitate such an approach due to their core values of democracy, mutual learning, and user empowerment.

3) The ethical dimensions of care robots should be addressed more extensively. The same applies to legal aspects, particularly in light of new or emerging regulatory frameworks such as the AIA and RMP proposals. Human rights, such as rights to non-discrimination, inclusion, accessibility, and health, should also receive greater attention in future projects, particularly to reduce the risk of marginalizing vulnerable groups or elderly people with low levels of digital literacy.

4) Future projects should address the semantics and utility of key concepts, such as autonomy, safety, personal integrity, and dignity, which may acquire new valences with the development of new types of care robots.

5) Finally, future research projects should cater to the development of standards and reference architectures that will enable future robots to communicate with each other, allowing for robot–robot interaction, human–robot–robot interaction, or similar. All in all, considering these aspects of social and assistive robots for elderly people will ensure better deployment and use of such robots, which in turn can lead to increased market share in the future.

## ACKNOWLEDGMENTS

This work is partially supported by the Research Council of Norway as a part of the Vulnerability in the Robot Society (VIROS) project, under grant agreement 28828, and the Multimodal Elderly Care System (MECS) project, under grant agreement 247697.

## AUTHOR'S CONTRIBUTIONS

Author 1 initiated the paper and participated in the study design, selecting research projects, writing the introduction and background, and analyzing, drafting, and revising the paper. Author 2 participated in the analysis, discussion, drafting, and revision of the paper. Author 3 participated in acquiring funding and drafting and revising the paper. Author 4 participated in the study design, in obtaining funding, and in drafting and revising the paper.